\documentclass[11pt]{article}
\pdfoutput=1

\usepackage[]{naacl2021}
\usepackage{times}
\usepackage{latexsym}
\usepackage[T1]{fontenc}
\usepackage[utf8]{inputenc}
\usepackage{booktabs} 
\usepackage{color}
\usepackage{amsmath}
\usepackage{amsfonts}
\usepackage{graphicx}
\usepackage{bbm}
\usepackage{multirow, multicol}
\usepackage{comment}
\usepackage{xspace}
\usepackage{upquote}
\usepackage{float}
\usepackage{alltt}
\usepackage{algpseudocode}
\usepackage{enumitem}
\usepackage{mathpartir}
\usepackage{wrapfig}
\usepackage{array}
\usepackage{todonotes}

\newcommand{\RUSS}[0]{\textsc{russ}\xspace}

\definecolor{comment-gray}{rgb}{0.5, 0.5, 0.5}

\definecolor{comment-red}{rgb}{0.8,0,0}

\title{Grounding Open-Domain Instructions to Automate Web Support Tasks}

\author{Nancy Xu \\ Computer Science Dept. \\  Stanford University \\ \small\texttt{xnancy@stanford.edu}  \And 
Sam Masling \\ Computer Science Dept. \\  Stanford University \\ \small\texttt{smasling@stanford.edu} \And
Michael Du \\ Computer Science Dept. \\  Stanford University \\ \small\texttt{mdu7@stanford.edu} \AND
Giovanni Campagna \\ Computer Science Dept. \\  Stanford University \\ \small\texttt{gcampagn@stanford.edu} \And 
Larry Heck \\ Viv Labs \\  Samsung Research \\ \small\texttt{larry.heck@ieee.org}
\And 
James Landay \\ Computer Science Dept. \\  Stanford University \\ \small\texttt{landay@stanford.edu} \And 
Monica S Lam \\ Computer Science Dept. \\  Stanford University \\ \small\texttt{lam@stanford.edu}
}

\begin{document}
\maketitle 
\begin{abstract}
Grounding natural language instructions on the web to perform previously unseen tasks enables accessibility and automation. 
We introduce a task and dataset to train AI agents from open-domain, step-by-step instructions originally written for people.
We build \RUSS (Rapid Universal Support Service) to tackle this problem. \RUSS consists of two models:
First, a BERT-LSTM  with pointers parses instructions to ThingTalk, a domain-specific language we design for grounding natural language on the web. 
Then, a grounding model retrieves the unique IDs of any webpage elements requested in ThingTalk. 
\RUSS may interact with the user through a dialogue (e.g. ask for an address) or execute a web operation (e.g. click a button) inside the web runtime. 
To augment training, we synthesize natural language instructions mapped to ThingTalk.
Our dataset consists of 80 different customer service problems from help websites, with a total of 741 step-by-step instructions and their corresponding actions. 
\RUSS achieves 76.7\text{\%} end-to-end accuracy predicting agent actions from single instructions. It outperforms state-of-the-art models that directly map instructions to actions without ThingTalk. Our user study shows that \RUSS is preferred by actual users over web navigation. 
\end{abstract}

\begin{figure}[h!]
  \includegraphics[width=\linewidth]{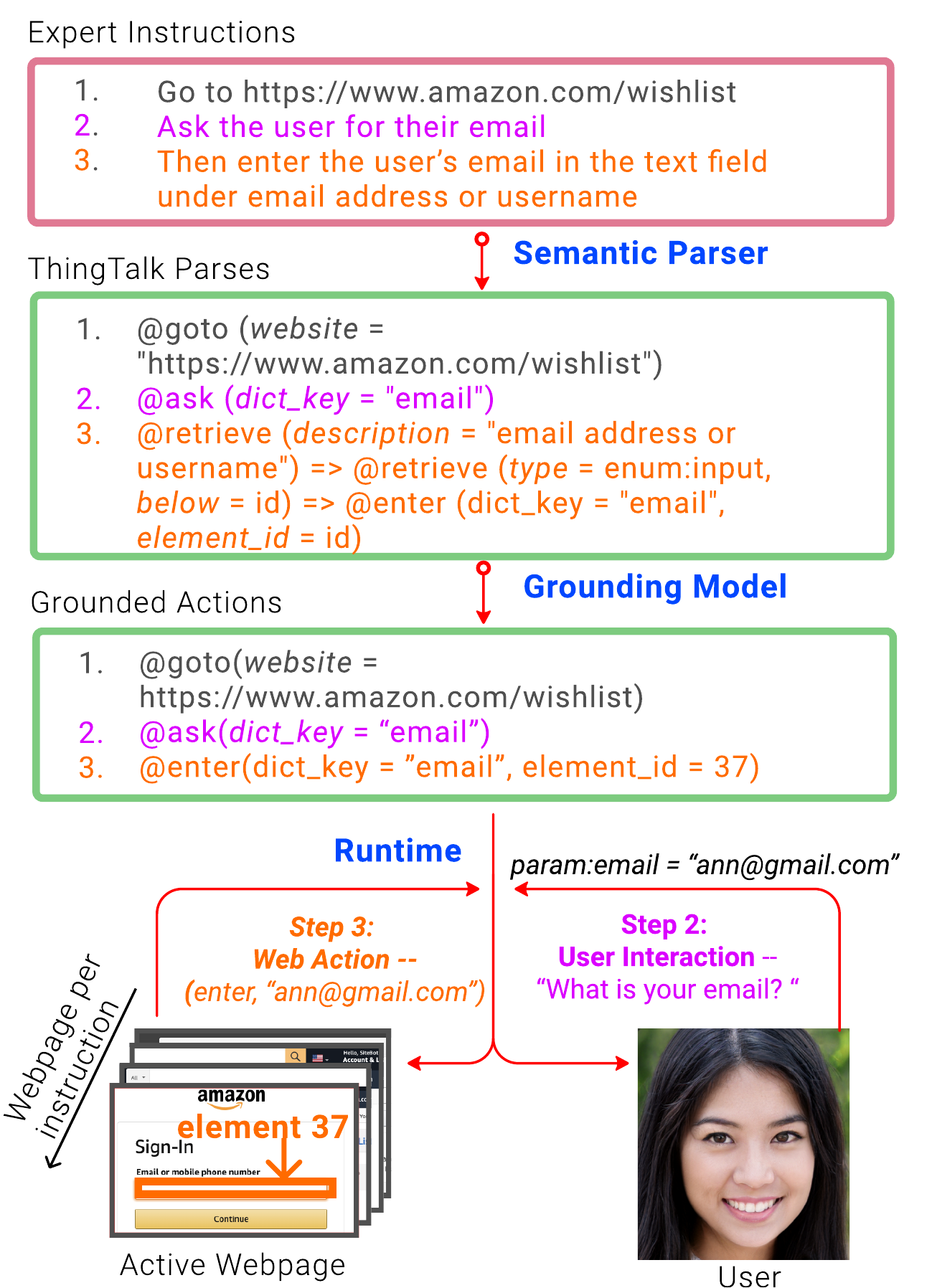}
  \caption{\RUSS 's semantic parser maps natural language instructions into ThingTalk (our DSL) and uses a grounding model to resolve elements in ThingTalk for grounded actions. The runtime executes the actions.}
  \label{fig:intro}
\end{figure}
\section{Introduction}
Grounding natural language is a key to building robots and AI agents \cite{chen:aaai11} that interact seamlessly with people. Besides grounding tasks visually \cite{mirowski2018learning, venugopalan:iccv15}, future AI agents must be able to ground language and execute actions on the web. 

We build a general-purpose, interactive agent to master tasks from open-domain natural language instructions on websites. 
We focus on the service domain for tasks such as redeeming a gift card, logging out of all your accounts, or resetting a password.

Conversational agents capable of providing universal access to  the web through a language interface are an important step towards achieving information equity. These agents empower those who are visually impaired or situationally preoccupied (e.g. driving) to obtain web-based knowledge and services for which they would otherwise require a laptop or mobile device for~\cite{sarsenbayeva2018situational}. Already, virtual assistants and call centers demonstrate a large number of scenarios where language interfaces backed by web backends are required by companies and users. However, unlike virtual assistants, web agents like \RUSS are universal, navigating the Web,  interacting with users, and bypassing the need for domain-specific APIs. 

On average over 60$\%$ of Americans have contacted customer service in a month \cite{servicestats}. A call center manager might instruct its agents to do the following to help a customer through a password reset: ``{\em go to passwordreset.com; ask the user for their desired new
password; click the reset button}''. As the agent performs the instructions on the web behind-the-scenes, the user is read information or asked questions periodically over a conversational interface (such as a phone).

Our approach, \RUSS (Figure \ref{fig:intro}), trains an agent that masters any web task specified from open-domain instructions. To do so, we design a domain-specific language (DSL) for grounding on the web and implement it as a subset of the ThingTalk programming language~\cite{genie}. Each natural language instruction maps to one of six agent actions that interact with users or operate on webpages. Actions that operate on the web are passed element IDs that are retrieved from high-level user language by grounding its corresponding ThingTalk on the active webpage. In the following, we use ThingTalk to refer to our subset taylored to web operations, where not ambiguous.
We break down the problem into two components: (1) a semantic parser that takes single-step natural language instructions and maps to ThingTalk statements using a BERT-LSTM pointer network, and (2) a grounding model that takes ThingTalk and retrieves an element ID on the active webpage where needed. 

The contributions of this work include:
\begin{enumerate}
\itemsep-0.2em 
\item \textbf{Task}: The new problem of building an interactive web agent capable of mastering tasks from open-domain natural language instructions. 
\item \textbf{\RUSS}: A fully functioning agent that services user support requests from natural language instructions. \RUSS consists of a semantic parser,
a grounding model, and a runtime. We release \RUSS as an open-source repository \footnote{https://github.com/xnancy/russ}
\item \textbf{ThingTalk}: A typed DSL that grounds natural language instructions on the web. ThingTalk is designed to be an expressive target for natural language semantic parsing, 
and amenable to training data synthesis.
\item \textbf{\RUSS Dataset}: a) Evaluation: a collection of 741 real-world step-by-step natural language instructions (raw and annotated) from the open web, and for each: its corresponding webpage DOM, ground-truth ThingTalk, and ground-truth actions; and b) Synthetic: 
a synthetic dataset of 1.5M natural language instructions mapped to ThingTalk.  
\item  \textbf{Evaluation of \RUSS}: 76.7$\%$ accuracy on our \RUSS evaluation dataset.  Our semantic parser maps natural language instructions to ThingTalk at 85\% accuracy and our grounding model achieves 75\% accuracy in resolving web element descriptions. A user study of \RUSS shows preference of the natural language interface over existing Web UIs. 
\end{enumerate}
 
\section{Related Work}

{\bf Grounding in the Visual and Physical Worlds (Robotics)}.
Grounding language in both the physical world \cite{chen:aaai11} and in images and videos (\cite{venugopalan:iccv15}, \cite{groundingvisual}) through systems like visual question-answering \cite{vqa} have been extensively explored. For example, \newcite{thomason2016} describe the game ``I Spy'' where human and robot take turns describing one object among several in a physical environment, requiring grounding of natural language to the physical world, and robot-human dialogues are explored in \cite{robothumandialog}. Previous work has proposed adaptive language interfaces for robots in dyanmic settings such as \cite{workflowrlweb}, \cite{chainingtriggeraction, workflowrlweb}, \cite{adaptiveinterfacedecomposition}, and \cite{arramon}. Other work builds physical world agents that operate through sequential actions \cite{chen:aaai11,misra2017mapping, mirowski2018learning}. 

\begin{table*}[h]
\small
\centering
\begin{tabular}{p{5.3cm}|p{9cm}}
\toprule
\textbf{Agent Action} & \textbf{Description} \\
\midrule
@goto(\textit{url}) & Navigate to the given URL \\ 
\hline
@enter(\textit{element\_id}, \textit{dict\_key}) & Find the closest match to the given dictionary key and enter its value in the given input element\\
\hline
@click(\textit{element\_id}) & Click on the given element \\ 
\hline
@read(\textit{element\_id}) & Read the content of the given element to the user \\ 
\hline
@say(\textit{message}) & Read the given message to the user \\
\hline
@ask(\textit{dict\_key}) & Ask the user for the value of a dictionary key \\
\bottomrule
\toprule
\textbf{Grounding Function} & \textbf{Description} \\
\midrule
\text{@retrieve}(\textit{descr}, \textit{type}, \textit{loc}, \textit{above}, \textit{below}, \textit{right\_of}, \textit{left\_of}) : \textit{element\_id} & Retrieves the elements matching the descriptors, returns an \textit{element\_id}.\\
\bottomrule
\end{tabular}
\caption{WebLang Agent Actions and a Grounding Function}
\label{tab:weblang}
\end{table*}
 
{\bf Natural Language Digital Interfaces}. An intelligent automated software assistant that collaborates with humans to complete tasks was first introduced in \cite{plow}. Since then, identifying UI components from natural language commands has been an important area of research in grounding, with prior work investigating approaches to map natural language instructions to mobile interfaces such as Android \cite{mapui} and Adobe photo editing GUIs \cite{editme}. Earlier work mapped natural language commands to web elements on a TV screen through a combination of lexical and gesture intent \cite{heck2013multi}.  More recently, \newcite{phrasenode} attempted to map natural language commands written by Amazon Mechanical Turkers to web elements (without actions). Unlike prior research, our work focuses on a new domain of parsing natural language instructions into \text{\it executable actions} on the web, where instead of mapping directly to elements using a neural model, we semantically parse natural language instructions to formal actions that support web navigation as well as user interactivity.

{\bf Dialogue Agents for The Web}. Other web-based dialogue agents are developed through single-use heuristics and more recently through programming-by-demonstration (PBD) tools. This approach allows users and developers to author programs that operate on the web and invoke those programs in natural language~\cite{sugilite, 10.1145/3210240.3210339, vash, geno}.
CoScripter~\cite{10.1145/1357054.1357323} additionally allows the user to edit the demonstration in natural language, and parses a limited natural language into executable form. While related in end goal, our work does not require user demonstration and can operate using existing real-world instructions. We note though that the WebLang intermediate representation and our grounding model can be used to improve the robustness of PBD systems as well.


\section{Task and Model}

Given a set of natural language instructions  $S = (i_1, \ldots, i_n)$ and a starting web page, our task is to construct an agent that follows the instructions through a series of action $A=(a_1, \ldots, a_n)$. Actions include web navigation and end-user interaction in order to obtain necessary information. 
Surveying online customer service tasks, $6$ action operations were identified as necessary for agents: open a URL page, enter text, click on buttons, say something to the user, read the results to the user, and ask user for some information. Details  are described in Table~\ref{tab:weblang}, where elements on a web page are assumed to be given unique element IDs.

\RUSS is trained to execute tasks by grounding natural language instructions on the web. The modular design of \RUSS, with separate semantic parser and grounding model,  is motivated by the high cost of training data acquisition, and the ability to improve each component independently.

We first describe ThingTalk, then the three components of Russ: the semantic parser model, the grounding model, and the runtime. 

\subsection{ThingTalk}
ThingTalk is designed to be (1) robust to open-domain natural language, (2) a suitable target for semantic parsing from natural language, 
and 
(3) trainable with only synthetic data.

The primitives in ThingTalk include all the agent actions and a grounding function @retrieve (Table~\ref{tab:weblang}).  The latter is informed by the descriptions in the instructions we found in the wild. The input features accepted by @retrieve are:
\begin{itemize}
\itemsep-0.5em 
  \item \textit{descr}: textual description of the element
    \item \textit{type}: type of element (button, input box, paragraph, header, etc.)
    \item \textit{loc}: absolute position of the element on the page
    \item \textit{above/below/...}: position of the element relative to another; above, below, right, and left.
\end{itemize}


To support element descriptions involving multiple features or other elements, ThingTalk is compositional in design. A ThingTalk program is a sequence of statements with syntax 
$\left[r \Rightarrow\right]^* a$, where $r$ is the retrieve operation and $a$ is an agent action. @retrieve returns an element\_id that is passed to @click (to click on the element), @read (to read the text in the element to the user), or @enter (to enter text in the element). For agent actions that require an element id, we call the sequence of @retrieve functions used to obtain the final element id used in the agent action the \textit{query}. See Figure \ref{fig:intro} for sample ThingTalk parses from natural language instructions. The orange ThingTalk parse demonstrates a query with 2 @retrieve functions. 

\subsection{Semantic Parser Model}
\label{sec:parser}
\begin{figure}
\centering
\includegraphics[width=\linewidth]{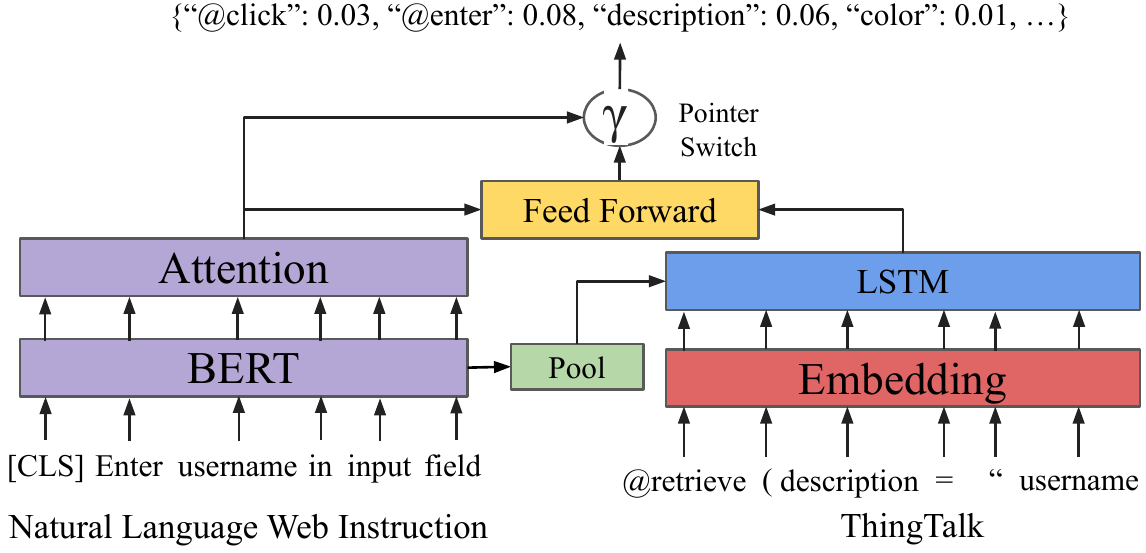}
\caption{The \RUSS semantic parser, using the BERT-LSTM architecture}
\label{fig:semanticparser}
\end{figure}
To translate natural language instructions into ThingTalk, we use the previously proposed BERT-LSTM model~\cite{schema2qa}. BERT-LSTM is an encoder-decoder network that uses a pre-trained BERT encoder~\cite{bert} and LSTM~\cite{hochreiter1997long} decoder with a pointer-generator~\cite{see2017get,paulus2018}. The architecture is shown in Fig.~\ref{fig:semanticparser}. The model is trained to encode natural language utterances and produce the ThingTalk code token-by-token. The pointer network in the decoder allows the model to predict out-of-vocabulary words by copying from the input utterances.

We preprocess the natural language by performing \textit{entity extraction}, where entity strings are mapped to placeholder tokens (URL, LOC, TYPE), and the strings are substituted back into the ThingTalk code after parsing with the placeholder tokens. This resolves errors related to long URLs being broken into tokens that are not always copied to ThingTalk together and helps disambiguate important input features. For example: "Click the button on the top of the amazon.com page" maps to "Click the TYPE on the LOC of the URL page". We use a simple set of heuristics to identify the entity strings for each placeholder token, such as the presence of a 'www.', '.com', 'http' substring to indicate a URL entity.

\subsection{Grounding Model}
The webpage is modeled using the Document Object Model (DOM), which is a hierarchical representation of all elements in the page.  Our DOM representation records {\em element features} such as the following for each element: 
\begin{itemize}
\itemsep-0.3em 
    \item inner text content of the element
    \item HTML \textit{id}, \textit{tag}, \textit{class}
    \item \textit{hidden} state (True/False if element is visible on the webpage)
    \item height/width of the element
    \item left/right/top/bottom coords of the element
    \item list of child elements in the DOM. 
\end{itemize}
An example is shown in Fig.~\ref{fig:dataset-example}.
\begin{figure}
\small
\begin{tabbing}
12\=12\=12\=\kill
\textbf{Instruction}: ``Enter the user's order number in the text\\
\>\>field that says order number'' \\
\textbf{DOM}:\\
\>element\_id: 1, type = "body"\\ 
\>\>element\_id: 2, type = "h1", text = "Your Orders"\\ 
\>\>element\_id: 3, type = "form"\\
\>\>\>$\ldots$\\
\>\>\>element\_id: 48, type = "label", text = "order number"\\
\>\>\>element\_id: 49, type = "input"\\
\>\>$\ldots$\\
\textbf{ThingTalk}:\\
\>$\text{@retrieve}(\textit{description} = \text{``order number''}, \textit{type}=\text{input})$\\
\>\>$\Rightarrow \text{@enter}(\textit{text}=\textit{order\_number}, \textit{element}=\textit{id})$\\
\textbf{Action}: $\text{@enter}(\textit{text}=\textit{order\_number}, \textit{element}=49)$
\end{tabbing}
\caption{Representation of an instruction in \RUSS}
\label{fig:dataset-example}
\end{figure}

\RUSS's grounding model grounds a ThingTalk @retrieve function by mapping it to an element ID. The input features in the @retrieve function are mapped against scores derived from the element features in the DOM to identify the best match.


The grounding model consists of the following steps. It filters elements by their type and absolute location. Next it handles relative positioning by identifying those elements with the right relational context to, and not too far away from, the given element's coordinates. It passes the text of the remaining candidates through a Sentence-BERT \cite{sentencebert} neural network and computes the cosine similarities of their embeddings with the embedding of the input text description.  The elements with the highest score are returned. 

\subsection{The Run-Time}
To execute the grounded ThingTalk program, \RUSS starts a new automated Chrome session for each task and uses Puppeteer to automate web actions in the browser. \RUSS uses a Google Voice API to implement actions involving user interactions (@say, @ask, or @read). For @ask actions, \RUSS uses a preprogrammed dialogue to ask the user for a dictionary key (such as ``name''), verifies the dictionary key is a valid string, and stores the value given by the user in a user's dictionary under that key. In @enter actions, we retrieve information to be entered by finding its closest match among the user's dictionary keys. 

\begin{table*}[h]
\centering
\small
\begin{tabular}{llc}
\hline
\textbf{ThingTalk Includes: (@retrieve feature)} & \textbf{Description}  & \textbf{Frequency} \\
\hline

Type reasoning (type) & Requires specific HTML type (e.g. button, checkbox) & 29.0\%\\
Input target (type = input)& Requires target element is a text input  & 25.0\% \\
Relational reasoning (below/above/left of...) & References neighboring features of the element & 10.3\% \\
Spatial reasoning (location) & References element location on the webpage& \hspace{0.15cm}4.6\% \\
No web element (No @retrieve) & No element (operation is @ask / @goto / @say)& 38.6\% \\
\hline
\end{tabular}
\caption{Subset of reasoning types (with the @retrieve input feature used to indicate it) supported by ThingTalk and their frequency in the \RUSS dataset. Some statements require multiple reasoning types.
}
\label{tab:reasonings-table}
\end{table*}

\section{Datasets}

This paper contributes two detasets, the \RUSS Evaluation Dataset with real-world instructions and the \RUSS Synthetic Dataset for training semantic parsers.

\subsection{\RUSS Evaluation Dataset}

The \RUSS Evaluation Dataset consists of real-world tasks from customer service help centers of popular online companies. To make our task-open domain, the online help centers we use span a diverse range of domains including music, email, online retail, software applications, and more. For each instruction in a task, the dataset includes:
\begin{itemize}
\itemsep-0.4em 
\item the English instruction in natural language as it appears in the original website, and the human-edited version of the instruction
\item the DOM of the web page where the instruction can be executed, with the element features associated with each element
\item the ThingTalk code corresponding to the instruction
\item the grounded action of the instruction
\end{itemize}

\begin{figure}[h]
  \includegraphics[width=\linewidth]{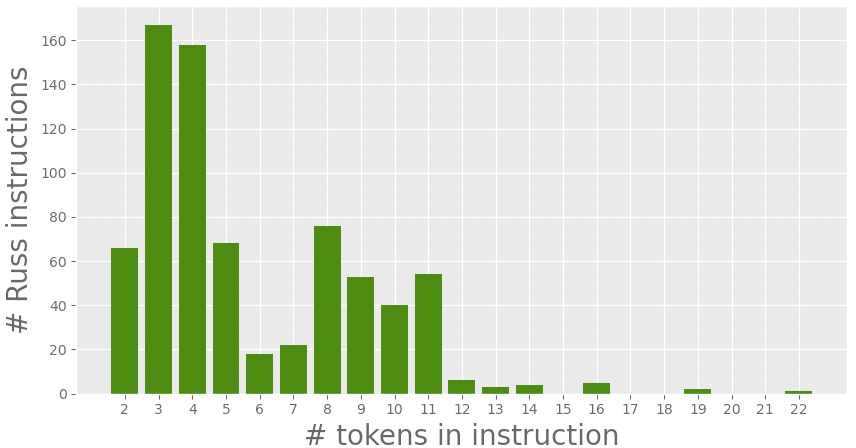}
  \caption{Lengths of instructions in the \RUSS Evaluation Dataset}
  \label{fig:token-distribution}
\end{figure}

\begin{figure}[h]
  \includegraphics[width=\linewidth]{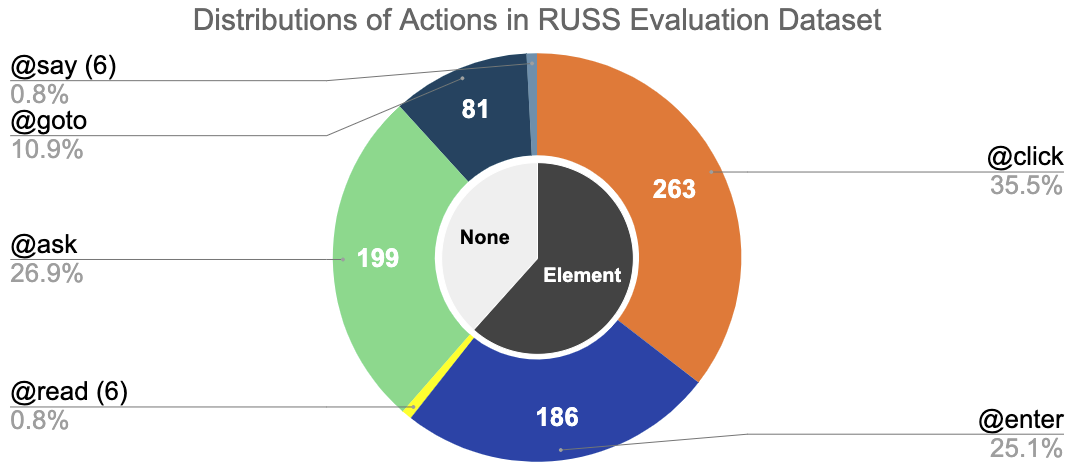}
  \caption{Distribution of actions in the \RUSS Evaluation Dataset. @click, @enter, and @read require a webpage element.}
  \label{fig:dataset}
\end{figure}

To collect the \RUSS Evaluation dataset, we acquire a list of ``Top 100 visited websites'' and locate tasks that offer line-by-line help instructions from those. An author of the paper walked through each task, performed the actions as instructed, scraped the webpage in the browser, and annotated the instruction with the corresponding ThingTalk code. Steps found missing from the instructions were inserted. If an instruction mapped to several actions, the text was broken into individual instructions. Note that the human worker did not participate in the design of ThingTalk; they were asked to write instructions as if they were teaching another human step-by-step.

\begin{figure}[h]
  \includegraphics[width=\linewidth]{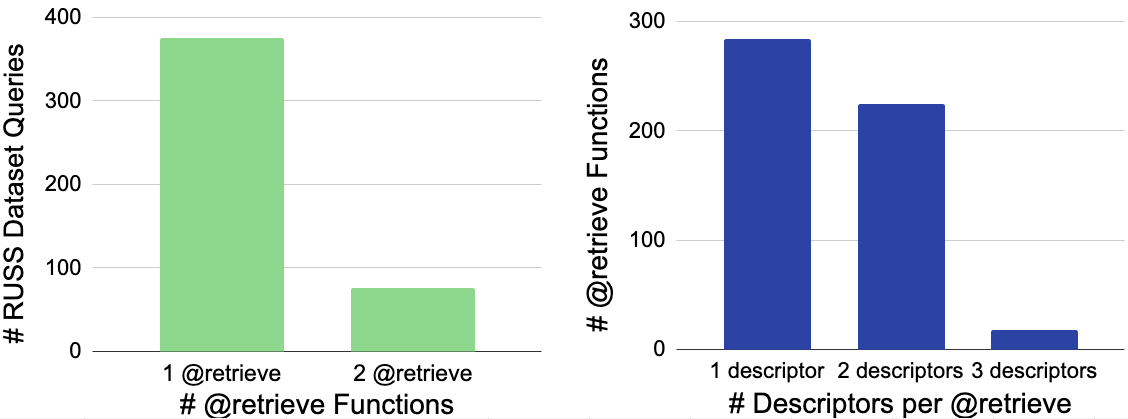}
  \caption{\# @retrieve functions in each \RUSS instruction and \# descriptors in each @retrieve.}
  \label{fig:descriptor-number}
\end{figure}

We collected a total of 80 tasks and 741 lines of instructions from 22 different online help centers.  The dataset is split into a dev set and a test set, with 304 instructions from 30 tasks in the dev set and 437 instructions from 50 tasks in the test set.  The \RUSS Evaluation dataset is not used for training.  On average, instructions in \RUSS contain 9.6 tokens (Fig.~\ref{fig:token-distribution}), significantly longer than the crowdsourced web instructions  in PhraseNode which average 4.1 tokens.
The three most common actions in the dataset are “click”, “ask” and “enter”  (Fig. \ref{fig:dataset}).
61.4\% of the natural-language instructions require retrieving an element from the webpage (click, enter, read). Table \ref{tab:reasonings-table} illustrates different types of reasoning supported by the @retrieve descriptors and their frequency in the \RUSS Evaluation Dataset. 
Lastly, 76 of the 455 element queries use two @retrieve functions, with the rest all just using one, and 53.7\%, 42.7\%, and 3.6\%  of the @retrieve functions have 1, 2, and 3 descriptors, respectively
 (Fig.~\ref{fig:descriptor-number}).

 While the language has just 7 core actions, the combinatorial space of possible actions and web elements is much larger -- on the order of 1000s of possible combinations per instruction. On average the DOMs of the webpages contain 689 web elements each. 

The total vocabulary size of the Evaluation Dataset found in the wild is 684 words. We find that at least one of the most frequent 300 words in the Evaluation vocabulary is present in >50\% of the Evaluation Dataset instructions. There are also many domain-specific words throughout the instructions.

\subsection{Synthetic Dataset}
Labeling large numbers of instructions in ThingTalk for training is time consuming and demands expertise. To address this, we use a typed template-based synthesis method to generate our training data. We write templates for each ThingTalk primitive and common combinations thereof. We also scrape a large dataset of naturally occurring DOM element text, webpage URLs, and phrases that are likely to be variable names to use for each parameter. The synthesizer compositionally expands the templates and sample values from the scraped dataset to construct a large training set of instructions mapped to ThingTalk automatically. We generate hundreds of different types of natural language templates which are combined to create a Synthetic Dataset with 1.5M training samples. This composition method creates roughly 840 distinct templates. To promote generalizability of our model, the total vocabulary size of the Synthetic corpus is large compared to the evaluation vocabulary size at 9305 words.

An example of a simple template is: 
\begin{center}
``At the \textbf{loc} of the page, \\
\textbf{@click} the button that says \textbf{descr}''
\end{center}
which is mapped to the ThingTalk: 
\begin{center}
@retrieve(\textit{descr} = \textbf{descr}, \textit{loc} = \textbf{loc}) $\longrightarrow$ \textbf{@click}(element = id)
\end{center}

\begin{table}
\small
\centering
\begin{tabular}{p{5cm}c}
\toprule
\textbf{Model} & \textbf{Accuracy (test)}\\
\midrule
\RUSS (1.5M training parses) & \bf 87.0\% \\
\toprule
\textbf{Ablations} & \textbf{Accuracy (dev)} \\
\midrule
\RUSS (1.5M training parses) & 88.2\% \\
$-$ entity extraction & 77.6\%\\ 
$-$ 1M training parses, entity extraction &70.0\%\\ 
\bottomrule
\end{tabular}
\caption{Evaluation of Semantic Parsing Model (trained on 1.5M parses) on \RUSS Evaluation test set. Ablations are performed on the dev set. ``$-$'' in Ablations subtracts a feature from the \RUSS model, the second ablation is trained on 500k training parses.}
\label{tab:parsereval}
\end{table}
\section{Evaluation}
\RUSS achieves \textbf{76.7\%} overall accuracy on the Evaluation Dataset, even though all of \RUSS, including the semantic parser is trained with only synthetic data. 

We perform 3 experiments to evaluate the individual components and the system as a whole:  1) Accuracy evaluation of \RUSS's Parsing Model with ablation studies. 2) Accuracy evaluation and baseline comparisons of \RUSS 's Grounding Model. 3) User study evaluating \RUSS's ability to master 5 tasks on-the-job. We test usability and efficacy of \RUSS compared with existing customer service help websites. 

\subsection{Semantic Parsing Accuracy}
Our first experiment evaluates the accuracy of our semantic parser on the \RUSS Evaluation dataset. 
We measure \textit{Exact Match Accuracy}: a parse is considered correct only if it matches the gold annotation token by token.

The results are shown in Table~\ref{tab:parsereval}. 
The parser obtains \textbf{87.0\%} accuracy on the test set. Despite using no real-world training data, the semantic parser achieves high accuracy on the challenging evaluation set.  It achieves an accuracy of 81.4\%  for instructions involving web elements, and 94.6\% for the rest. This suggests the semantic parser can handle both types of instructions with high accuracy, especially instructions that parse to user interactions (no web element). 

We perform an ablation study on the \RUSS Evaluation dev set as seen in Table \ref{tab:parsereval}. \RUSS achieves 88.2\% accuracy on the dev set.  The entity extraction technique where string entities are replaced with placeholders during training, as discussed in Section~\ref{sec:parser}, contributes 10.6\% improvement in accuracy. Training without this pre-processing step and with only 500K parses will reduce the accuracy further by 7.6\%.  This suggests that it is important to have a large synthetic training data set.

\begin{table}
\centering
\small
\begin{tabular}{lc}
\toprule
\textbf{Model} & \textbf{Grounding Acc (test)}\\
\midrule
\RUSS  &  \text{\bf 63.6\%} \\
End-to-End Baseline & \text{51.1\%} \\
PhraseNode & \text{46.5\%} \\
\bottomrule
\end{tabular}
\caption{\RUSS outperforms state-of-the-art PhraseNode in the grounding subtask on the \RUSS Evaluation test set.}
\label{tab:grounding}
\end{table}

\subsection{Grounding Evaluation}
With an effective semantic parser to ThingTalk, we next measure the grounding accuracy: the percent of correctly identified element\_ids from the 252 natural language commands referring to web elements in the \RUSS test set. As shown in Table \ref{tab:grounding}, \RUSS achieves an accuracy of 63.6\%. 81.4\% of the instructions are parsed correctly, and 77.9\% of the correct parses are grounded accurately.  Had the semantic parser been correct 100\% of the time, the Grounding Model would achieve an accuracy of 73.0\%. The semantic parser is more likely to correctly parse simple instructions such as "click sign in", which are also generally easier for the Grounding Model, explaining the delta between 77.9\% and 73.0\%.

We create an {\em End-to-end Baseline} model to compare against the 2-step approach of \RUSS. 
Here, we represent web elements using \RUSS's feature elements as before.  However, we do not parse the natural language sentences into their input features in \RUSS, but is left intact as input to Sentence-Bert to compute its embedding. Like Section 4.3, the element sharing the closest embedding with the input sentence is returned. This end-to-end baseline model performs with $12.6\%$ less accuracy than \RUSS, illustrating the benefits of using a semantic parser.

To compare our grounding model with state-of-the-art results, we also replicate the best performing embedding model from \cite{phrasenode}, which we reference as PhraseNode.  The webpage features used as inputs in PhraseNode are a subset of our representation. PhraseNode achieves an accuracy of 46.5\%, which is 4.6\% worse than our Baseline and 17.2\% lower than \RUSS.  We show that the combination of a high-performance semantic parser and a well-tuned grounding model can outperform the best end-to-end neural models for grounding on the web.  

\subsection{Analysis}
The entire one-time process for training \RUSS takes approximately 7 hours on an NVIDIA Tesla V100.  \RUSS can perform a new task on-the-job by running the instructions through the semantic parser in less than 1 minute.

We analyze how well \RUSS and PhraseNode perform for sentences in the Evaluation Set requiring different types of reasoning (Table~\ref{tab:groudingreason}). Russ outperforms the state-of-the-art PhraseNode~\cite{phrasenode} for all the reasoning types.  It performs well on grounding tasks that involve type, input, and relational reasoning. Evaluation of the spatial reasoning instructions revealed that many referenced image features (e.g. ``click the hamburger menu icon''), which is not supported by \RUSS.  The results show that ThingTalk is simple enough to be generated by a neural language model, while comprehensive enough to express the wide range of open-domain natural language instructions for web tasks. 

\begin{table}
\centering
\small
\begin{tabular}{lrrr}
\toprule
\textbf{Reasoning} & \bf \RUSS &  PhraseNode \\
\midrule
Type  & 67.8\% & 61.5\% \\
Input   & 75.6\% & 60.4\% \\
Relational   & 70.0\% & 53.5\%  \\
Spatial  & 36.7\% & 30.3\% \\
\bottomrule
\end{tabular}
\caption{Grounding Accuracy Comparison of \RUSS and PhraseNode by Reasoning type on the \RUSS Evaluation test set.}
\label{tab:groudingreason}
\end{table}

Unlike end-to-end models that struggle with long, complex instructions, we find that \RUSS benefits from added reasoning in instructions that constrains the potential set of element candidates (e.g. ``the element must be an input''). Webpages commonly have thousands of elements and the probability of matching the right element increases with constraints.

Of the $741$ instructions in the RUSS dataset, $6$ contain attributes that are not well expressed in ThingTalk. For example, ``select the user's birth month in the month drop down'' is not parsed correctly because ThingTalk does not have a notion of selecting an element in a menu. This feature will be added in the future. 


Another source of errors lies in how webpages are constructed.  Important attributes needed for grounding can be hidden behind classes.  For example, an element may be labeled as ``Click here'', but the text is not present in the DOM text attribute and instead obscured behind a site-specific class name such as "next-page-button". Grounding techniques on visual data can be helpful in resolving this class of problems. 


\subsection{User Study}
The goal of our user study is to evaluate the end-to-end feasibility of \RUSS on open-domain instructions from real customer service websites, and evaluate how users respond to \RUSS. This is a small-scale study with promising early results, but can benefit from further user studies on larger populations. 

We recruited 12 participants who were asked to complete 5 customer-support tasks (Table~\ref{tab:exstudy}), chosen from popular websites: Amazon, Spotify, pinterest, Google, and Walmart, with both \RUSS and the browser. For all tasks, users were given a fake persona (a set of credentials such as email, password, gift card code, etc) to use when interacting with the agent.  The study was approved by our IRB and participants were compensated. 

\begin{table}
\centering
\small
\begin{tabular}{c l}
\toprule
\textbf{\# 1} & Redeem Amazon Gift Card \\
\textbf{\# 2} & Get Pinterest Ad Account Number \\
\textbf{\# 3} & Log out of all Spotify accounts\\
\textbf{\# 4} & Create new Walmart account \\ 
\textbf{\# 5} & Send Google feedback \\
\bottomrule
\end{tabular}
\caption{Tasks in \RUSS User Study}
\label{tab:exstudy}
\end{table}

The participants in our study ranged from ages 21 to 68 years old, with an average age of 36 years old, a 50/50 male/female ratio, and varied technical sophistication. To reduce learning effects, we used Latin Square Balancing \cite{latin} to ensure that both the web and \RUSS trials of each site were performed first half the time. We record users' time to perform each task, number of turns (in \RUSS) or clicks (on the web) required to achieve each task, and gave each participant an exit survey containing qualitative assessments. 

\begin{figure}
  \includegraphics[width=\linewidth]{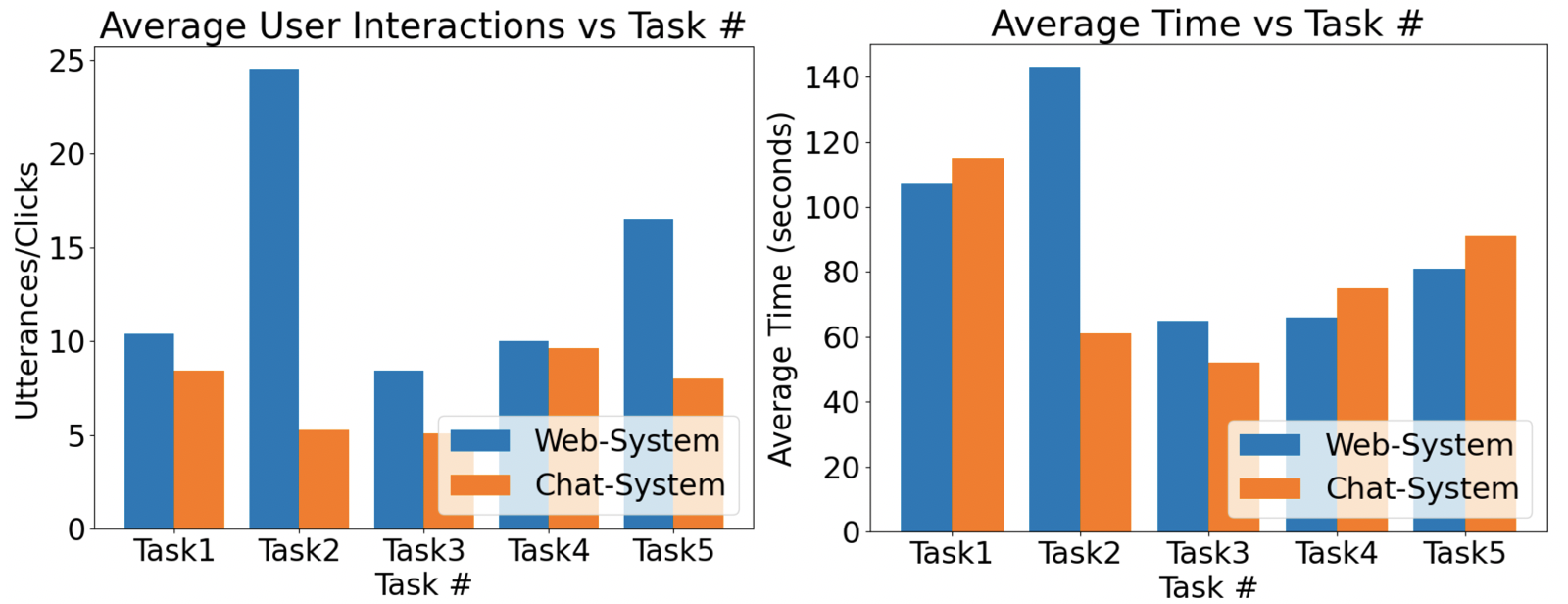}
  \caption{Average number of user interactions via utterance or click (left); average time taken to complete tasks in seconds (left)}
  \label{fig:quant}
\end{figure}

Participants were able to complete \textbf{85\%} of the tasks on their own on the web and \textbf{98\%} of tasks with the help of \RUSS. Those who did not finish their task either gave up or failed to complete the task within 5 minutes. 
The time it took users to accomplish each task was similar for the Web and \RUSS  (Fig.~\ref{fig:quant}), though \RUSS was significantly faster for Task 2, a more complex task users said they were unfamiliar with. This seems to indicate that \RUSS is more favorable for unfamiliar, complex tasks.

\begin{figure}
  \includegraphics[width=\linewidth]{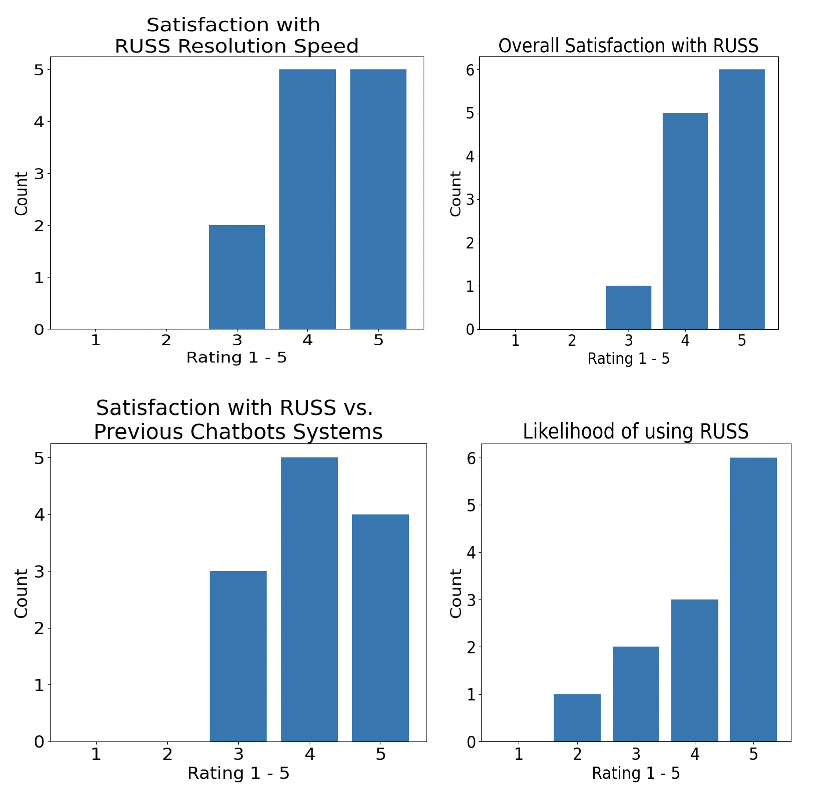}
  \vspace{-0.6cm}
  \caption{Qualitative results from user studies. On a scale 1-5 for satisfaction, 1 = not satisfied at all and 5 = exceeded expectations. For likelihood, 1 = will never use again and 5 = will definitely use again. }
  \label{fig:qual}
\end{figure}

After trying the $5$ tasks, \textbf{$\bf 69\%$} of users reported they prefer \RUSS over navigating online help pages. Reasons cited include  ease of use, efficiency, and speed, even though the times of completion were similar. Participants were generally pleased with their \RUSS experience, and only one person said that they were unlikely to use \RUSS again (Fig.~\ref{fig:qual}). However, many users did report that they wished \RUSS was as visually stimulating as the browser. Other users noted that they felt more familiar and comfortable with the browser. 

As a final discussion, it is worth noting that while the user study results are extremely promising, this is a small scale study. \RUSS 's runtime needs stronger error handling for out-of-context conversation. Currently, \RUSS gives the user 3 tries to return an expected response before terminating. \RUSS also times out if a webpage takes more than >60 seconds to load in Puppeteer. We saw instances of both of these situations in the \RUSS user study in the few cases the user failed to complete a task. 

\section{Conclusion}
 \RUSS demonstrates how a semantic parser and grounding model can be used to perform unseen web tasks from natural language instructions. By achieving 76.7\% accuracy on the \RUSS Evaluation Dataset, we show how a modular semantic parsing approach can outperform end-to-end neural models on this task, and demonstrate how humans interact with \RUSS -like systems in the user study. Like many datasets in NLP, we believe extensive research is still required to go from RUSS’s 76.6\% overall accuracy on the Evaluation Dataset to 100\%. As seen in Table 4, prior models like PhraseNode achieve only 46.5\% grounding accuracy, which points to additional work necessary in grounding natural language on the web.

 The \RUSS Evaluation dataset introduces a set of real instructions for grounding language to executable actions on the web to evaluate future research in this direction, including training semantic parsers to new targets using real-world instructions and neural models for grounding formal language representations on the web. Our work provides the task, technical foundation, and user research for developing open-domain web agents like \RUSS. 
\section{Ethical Considerations}
The user study conducted in this paper was submitted to the Institutional Review Board and received IRB Exempt status. All participants were read an IRB consent form prior to the user study, which detailed the study details. No deception was involved in the study: all participants knew they were evaluating an AI agent in the conversation portion of the user study and were not led to believe otherwise. They study took about 20 minutes. All participants were compensated with \$10. 

The webpages scraped for the \RUSS dataset are all public domain webpages. No individual personal identifying information was used to obtain the webpages. On websites that required accounts to access pages, we created fake user accounts with non-identifying usernames / passwords / emails to navigate the websites in order to limit any privacy risks that may be involved. 

In the future, we see web agents like \RUSS helping improve accessibility by helping individuals who are visually impaired, less technologically advance, or otherwise preoccupied receive equitable access to information. Before systems like \RUSS are put to practice at scale, the authors believe more research must be done in understanding user behavior with web agents to safeguard against downstream consequences of system errors and to better understand how information can be effectively delivered by AI agents that operate in potentially high-stakes transactions such as health or finance. Our user study is the first step in this direction.  


\section{Acknowledgments}
 We thank Silei Xu for helpful discussions on constructing the Synthetic dataset, and Richard Socher for feedback and review of the final publication. 
 
 This work is supported in part by the National Science Foundation under Grant No. 1900638 and the Alfred P. Sloan Foundation under Grant No. G-2020-13938.

 Any opinions, findings, and conclusions or recommendations expressed in this material are those of the authors and do not necessarily reflect the views, policies, or
endorsements of outside organizations. 

\bibliography{anthology,custom}
\bibliographystyle{acl_natbib}


\end{document}